\pdfoutput=1
\documentclass[11pt]{article}

\usepackage[final]{acl}

\usepackage{times}
\usepackage{latexsym}

\usepackage[T1]{fontenc}

\usepackage[utf8]{inputenc}

\usepackage{microtype}

\usepackage{inconsolata}

\usepackage{graphicx}

\newcommand{\ours}{LightThinker}
\usepackage{xcolor}
\usepackage{amsmath}
\usepackage{multirow}
\usepackage{bbm}
\usepackage{booktabs}
\usepackage{colortbl}
\usepackage{tcolorbox}
\usepackage{amsfonts}

\definecolor{mygray}{gray}{0.9} 
\definecolor{myblue}{HTML}{F0FFFF}

\usepackage{calc}
\newlength\myheight
\newlength\mydepth
\settototalheight\myheight{Xygp}
\title{LightThinker: Thinking Step-by-Step Compression}

\author{
  Jintian Zhang${^{\spadesuit\heartsuit}\footnotemark[1]}$~, 
  Yuqi Zhu$^{\spadesuit\heartsuit}$\thanks{$\quad$ Equal Contribution.}~,
  Mengshu Sun$^{\clubsuit\heartsuit}$, 
  \textbf{Yujie Luo}$^{\spadesuit\heartsuit}$,  \\ 
  \textbf{Shuofei Qiao}$^{\spadesuit\heartsuit}$, 
  \textbf{Lun Du}$^{\clubsuit\heartsuit}$, 
  \textbf{Da Zheng}$^{\clubsuit\heartsuit}$, 
  \textbf{Huajun Chen}$^{\spadesuit\heartsuit}$, 
  \textbf{Ningyu Zhang}$^{\spadesuit\heartsuit}$\thanks{$\quad$ Corresponding Author.} \\
  $^\spadesuit$Zhejiang University ~$^\clubsuit$Ant Group \\ 
  $^\heartsuit$Zhejiang University - Ant Group Joint Laboratory of Knowledge Graph 
  \\
  \texttt{\{zhangjintian,zhangningyu\}@zju.edu.cn} \\
  \raisebox{-\mydepth}{\includegraphics[height=1.6\myheight]{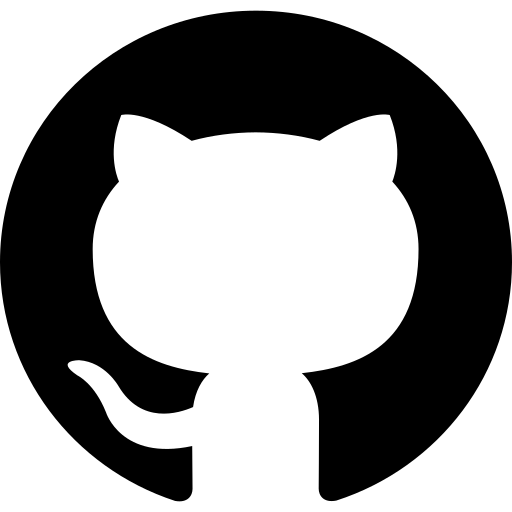}}
\textbf{\url{https://github.com/zjunlp/LightThinker}}
}

\begin{document}
\maketitle
\begin{abstract}
Large language models (LLMs) have shown remarkable performance in complex reasoning tasks, but their efficiency is hindered by the substantial memory and computational costs associated with generating lengthy tokens. In this paper, we propose \textbf{\ours}, a novel method that enables LLMs to dynamically compress intermediate thoughts during reasoning. Inspired by human cognitive processes, \ours~compresses verbose thought steps into compact representations and discards the original reasoning chains, thereby significantly reducing the number of tokens stored in the context window. This is achieved by training the model on when and how to perform compression through data construction, mapping hidden states to condensed gist tokens, and creating specialized attention masks. Additionally, we introduce the {Dependency (Dep)} metric to quantify the degree of compression by measuring the reliance on historical tokens during generation. Extensive experiments on four datasets and two models show that \ours~reduces peak memory usage and inference time, while maintaining competitive accuracy. Our work provides a new direction for improving the efficiency of LLMs in complex reasoning tasks without sacrificing performance. 
\end{abstract}


\section{Introduction}

\begin{figure}[!t] 
    \centering
    \scalebox{0.8}{
    \includegraphics[width=1\linewidth]{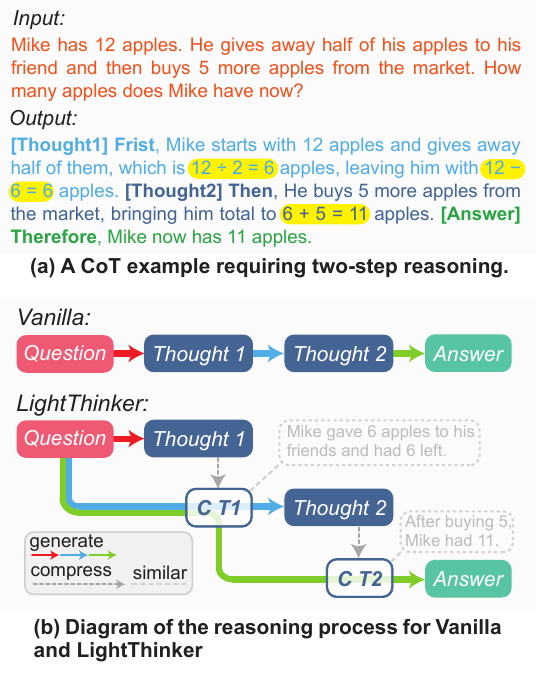} 
    }
    \caption{
    (a) A CoT case. Tokens highlighted in yellow represent critical reasoning tokens, while the remaining tokens primarily ensure fluency. 
    Humans typically only write the yellow parts when solving this problem.
    (b) Comparison of reasoning between Vanilla and \ours. 
    ``C Ti'' denotes the i-th compressed thought representation, and we illustrate the semantic information potentially expressed after compression.
    }
    \label{fig:intro}
    \vspace{-4mm}
\end{figure} 

Recent advancements in Large Language Models (LLMs) have demonstrated their remarkable capabilities in complex reasoning tasks \cite{zhao2023survey,azaria2024chat}. 
As research in this domain progresses, the reasoning patterns of these models have gradually evolved from ``fast thinking'' to ``slow thinking''. 
This transition is exemplified by methods such as Chain-of-Thought (CoT)~\citep{nips22_cot} prompting, which enhances reasoning by breaking down complex problems into sequential sub-steps.  
Building on this, the \textit{o1-like thinking mode}~\citep{arixv24_o1,arxiv24_qwq,arxiv25_deepseek_r1} introduces multiple reasoning abilities such as trial-and-error, backtracking, correction, and iteration, further improving the success rate of models in solving complex problems. 
However, this performance improvement comes at the cost of generating a large number of tokens~\citep{arxiv24_o1_study}. 
Given that current LLMs are predominantly based on the Transformer architecture~\citep{nips17_transformer}, the computational complexity of the attention mechanism grows quadratically with the context length, while the storage overhead of the KV Cache increases linearly with the context length.
For example, in the case of Qwen32B~\citep{arxiv24_qwen2_5}, when the context length reaches \(10^4\), the KV Cache occupies a space comparable to the model itself. 
Consequently, the increase in token generation leads to a sharp rise in memory overhead and computational costs, severely limiting the practical efficiency of LLMs in long-text generation and complex reasoning tasks.

To mitigate this issue, two main approaches have been proposed, primarily differentiated by their intervention requirements during inference.
The first category requires no additional intervention during inference, achieving efficiency through prompt engineering~\citep{arxiv24_tale,arxiv24_break_the_chain,arxiv24_concise_thoughts} or specialized training~\citep{nips24_skip_steps,arxiv24_c3ot,arxiv25_related_work_rl1,arxiv25_o1_pruner,arxiv24_ccot,arxiv24_coconut} to guide LLMs in generating fewer or even zero~\citep{arxiv23_kd_cot,arxiv24_icot} intermediate tokens during reasoning. 
The second category operates through real-time token-by-token intervention during inference~\citep{nips23_h2o,arxiv24_sepllm}, reducing memory usage by selectively retaining important parts of the KV Cache while discarding less critical ones.
However, both approaches face distinct challenges: the first typically requires careful data construction and iterative refinement, while the second introduces substantial inference latency due to the computational overhead of token-wise importance assessment.

In this work, we propose a new approach by training LLMs to dynamically compress historical content during reasoning. 
Our motivation stems from two observations: 
1) Tokens generated by the LLM serve dual purposes: 
ensuring linguistic fluency and facilitating actual reasoning (as highlighted in yellow in Fig.~\ref{fig:intro}(a)), making compression feasible. 
2) When humans solve problems similar to the one in Fig.~\ref{fig:intro}(a), they typically write only key steps (highlighted in yellow), while storing the rest of the thought process mentally.

Based on these insights, we introduce \ours, a method that dynamically compresses intermediate thoughts during generation. 
As illustrated in Fig.~\ref{fig:intro}(b), after generating a lengthy thought step (e.g., \texttt{Thought i}), it is compressed into a compact representation (e.g., \texttt{C Ti}), and the original thought chain is discarded, with reasoning continuing based solely on the compressed content. 
This approach significantly reduces the number of tokens stored in the context window, thereby lowering memory overhead and computational costs.

\begin{figure*}[t]
    \centering
    \begin{minipage}[t]{0.73\textwidth}
        \vspace{0pt}
        \scalebox{1}{\includegraphics[width=\linewidth]{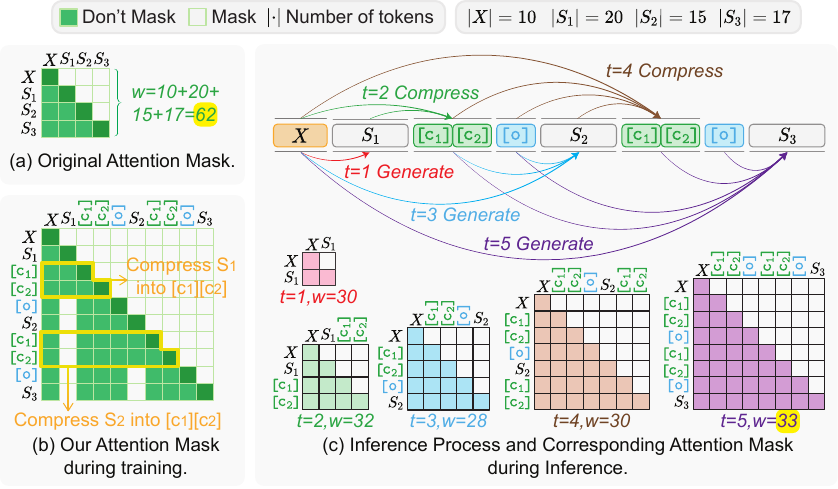}} 
        \caption{
        Overview of \ours, illustrated with an example requiring three-step reasoning.
        Fig. (a) shows the attention mask of Vanilla during both training and inference.
        Fig. (b) depicts the attention mask of \ours~during the training.
        Fig. (c) presents the complete inference process of \ours~along with the attention mask corresponding to each step. 
        Here, `w' denotes the size of the matrix.}
        \label{fig:method}
    \end{minipage}
    \hfill
    \begin{minipage}[t]{0.24\textwidth}
        \vspace{0pt}
        \scalebox{1}{\includegraphics[width=\linewidth]{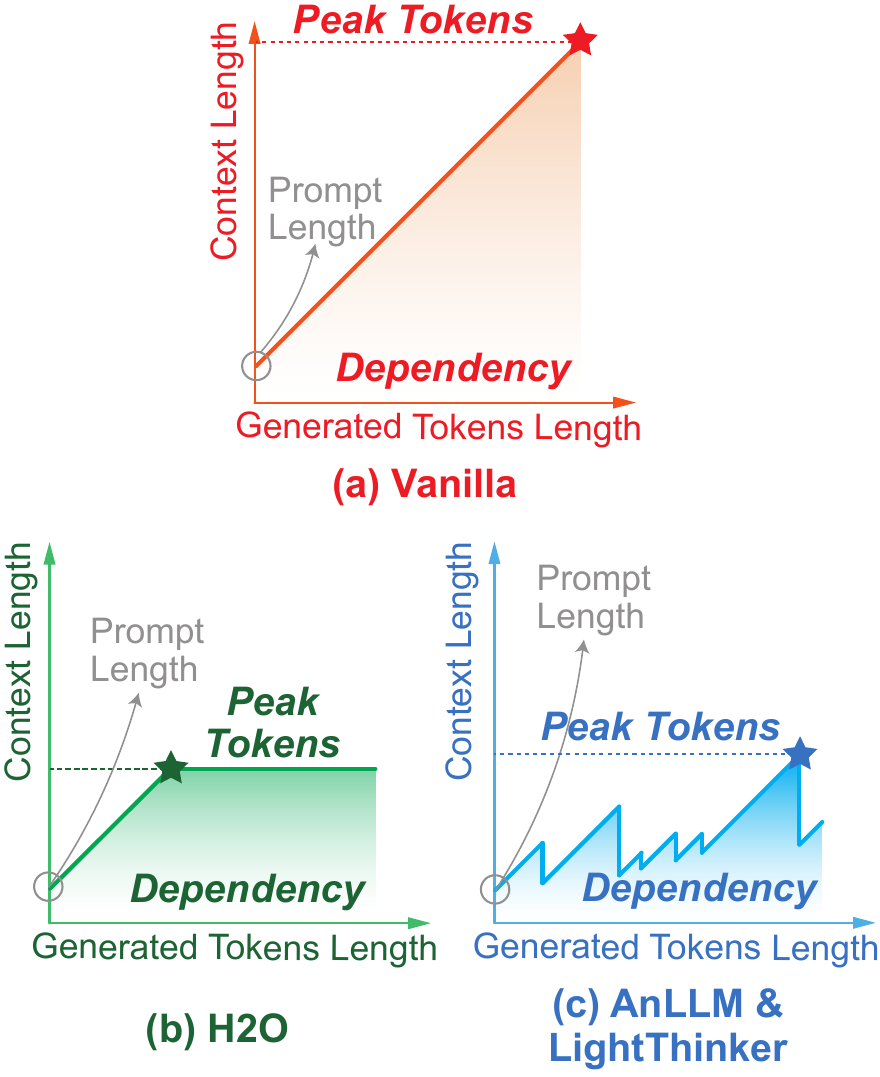}} 
        \caption{
        The relationship between context length and the number of generated tokens across different methods. 
        The \textit{Dependency} metric represents the area under the curve, while the \textit{Peak Token} denotes the maximum value of the curve.
        See Appx.~\ref{sec:app:metric} for details.
        }
        \label{fig:exp:metric}
    \end{minipage}
\end{figure*}

In practice, we train the LLM to learn when and how to compress. 
Specifically, we construct data to teach the model when to compress; the hidden states of the thoughts to be compressed are reduced to a set of hidden states corresponding to a small number of special tokens (i.e., gist tokens~\citep{nips23_gist}).
Through carefully designed attention masks, the LLM then learns how to compress and how to continue generating based on the compressed content.
{To quantify the amount of information used during reasoning, we further propose the Dependency (Dep) metric, defined as the total number of historical tokens each generated token depends on (see Fig.~\ref{fig:exp:metric}). 
This metric effectively measures the degree of compression, with a lower Dep value indicating reduced reliance on the original long context and more significant compression.}

We conduct extensive experiments across four datasets using two different models. 
The results indicate that, with the Qwen model, \ours~reduces the peak token usage by 70\% and decreases inference time by 26\% compared to the Vanilla model, while maintaining comparable accuracy (with only a 1\% drop).
Our contributions are as follows:
1) We propose \ours, a method that dynamically compresses thought chains during reasoning, significantly reducing memory overhead and inference time.
2) We introduce the Dependency (Dep) metric to measure the compression ratio and the amount of information used during reasoning.
3) We demonstrate that \ours~achieves a good balance between reasoning efficiency and accuracy, offering new insights for future LLM inference acceleration.


\section{Background}

\textbf{Slow Thinking.}
The reasoning ability of LLMs is crucial~\citep{acl23_reason_survey}, especially in solving complex problems, necessitating a shift from the fast-thinking System 1 to the slow-thinking System 2~\citep{pb96_system12,fsg11_thinking_fast_slow,aaai21_machine_fast_slow}. 
For instance, Chain-of-Thought (CoT)~\citep{nips22_cot} approaches decompose complex problems into sub-problems and solve them step-by-step. 
\textit{o1-like thinking mode}~\citep{arixv24_o1,arxiv24_qwq,arxiv25_deepseek_r1} goes a step further by incorporating abilities such as trial, reflection, backtracking, and correction on top of the divide-and-conquer strategy. 
Empirical evidence~\citep{arixv24_o1,arxiv25_deepseek_r1} shows that the \textit{o1-like thinking mode} significantly enhances the model's ability to solve complex problems compared to CoT. 
This slow-thinking mode can be instilled in models through carefully constructed data using Supervised Fine-Tuning (SFT). 
In terms of the number of output tokens, the token consumption of System 1, CoT, and \textit{o1-like thinking mode} increases progressively.

\paragraph{Inference Challenges.}
Recent works on \textit{o1-like thinking mode}~\citep{arxiv24_o1_study} 
highlight the necessity of generating a substantial number of tokens for complex problem-solving.
As the core structure of Transformers~\citep{nips17_transformer}, the attention mechanism faces two significant challenges during inference as token generation scales:
1) The \textit{memory} overhead gradually increases. 
To speed up inference, each token's Key and Value are cached at every layer. 
For the Qwen-32B~\citep{arxiv24_qwen2_5}, when the context length reaches $10^4$ tokens, the space occupied by the KV cache is comparable to that of the model itself.
2) The \textit{computational cost} of generating a single token in an autoregressive manner also increases. 
Due to the attention mechanism in Transformers~\citep{nips17_transformer}, the computational load grows \textit{quadratically} with the number of tokens.

\section{Methodology}
We propose \textbf{\ours} to accelerate the reasoning process of LLMs, as illustrated in Figure~\ref{fig:method}.
The core idea is to \textit{train LLMs to dynamically compress the current thought during reasoning}, enabling subsequent generation to be based on the compressed content rather than the original long thought.

\subsection{Overview}

\paragraph{Notation.}
For clarity, we define the following notations.
Lowercase letters, such as $x_i$, denote a single token.
Uppercase letters, such as $X$, denote sequences of tokens.
The notation `$\texttt{[·]}$' denotes a special token, such as `$\texttt{[c]}$', while `$\texttt{<·>}$' denotes an optional special token, such as `$\texttt{<w>}$'.
The \textit{o1-like thinking mode} dataset $\mathcal{D}=\{(X,Y)_i\}_{i=1}^{|\mathcal{D}|}$ consists of $|\mathcal{D}|$ samples, where $X=\{x_i\}_{i=1}^{|X|}$ represents a question, and $Y=\{y_i\}_{i=1}^{|Y|}$ represents the corresponding thought and final answer. 
Recent works~\citep{arxiv25_openthoughts,arxiv25_deepseek_r1} show that SFT on $\mathcal{D}$ significantly enhances LLM reasoning capabilities.

\paragraph{Design.}
To achieve the core idea, we focus on addressing two key questions:
\underline{\textit{i) When to compress?}} 
The timing of compression significantly impacts reasoning efficiency and compression quality. 
We explore two different strategies.
The first is \textit{token-level}~\citep{iclr25_activation_beacon}, where compression is performed after a fixed number of tokens.
This strategy is straightforward to implement but may ignore semantic boundaries.
The second is \textit{thought-level}~\citep{acl24_anllm}, where compression is performed after a complete ``thought'', defined as a sentence or paragraph.
This strategy better preserves semantic information but requires a more complex segmentation function.
\underline{\textit{ii) How to compress?}}
The goal of compression is to encode the current lengthy thought into a more compact representation.
We investigate two different approaches.
The first is \textit{text compression}, where the current thought is encoded into a shorter text~\citep{emnlp23_llmlingua} or a chunk of continuous vectors~\citep{emnlp23_autocompressors,iclr24_icae}.
This method requires an additional encoding model and increases computational overhead.
The second is \textit{hidden state compression}, where the hidden state of the current thought is compressed into the hidden states of a few special tokens (i.e., \textit{gist tokens}~\citep{nips23_gist}).
We choose this method as it does not require additional models.
Specifically, in our work, we address the first question by \textit{reconstructing data} and the second by \textit{constructing thought-based attention mask}.

\paragraph{What content has been compressed?}
We do not aim to compress lengthy thought information into a compact representation without loss. 
Instead, our focus is on preserving only the information that is essential for subsequent reasoning.
As highlighted by the gray dashed box in Figure~\ref{fig:intro}(b), the lengthy thought is retained solely for the elements that contribute to further inference.

\subsection{\ours}
\label{sec:method:method}

\textbf{Data Reconstruction.}
To enable LLMs to dynamically compress during generation, we reconstruct the original dataset $\mathcal{D}$ as follows.
First, we segment the output.
Given the input $X$ and output $Y$, we use a segmentation function $\texttt{Seg()}$ to divide $Y$ into $k$ subsequences $S$, i.e., $Y=\{S_i\}_{i=1}^{k}$.
The function can be based on \textit{token-level} or \textit{thought-level}.
Then, we insert the special tokens.
Specifically, we insert a set of special tokens $\{\texttt{<w>}, C, \texttt{[o]}\}$ between adjacent subsequences $S_i$, where $\texttt{<w>}$ is an \textit{optional} compression trigger, indicating the need to compress the preceding thought.
It can be omitted if the $\texttt{Seg()}$ is token-level or if $\texttt{<w>}\in S_i$.
The token $C=\{\texttt{[c}_\texttt{i}\texttt{]}\}_{i=1}^{|C|}$ consists of $|C|$ special tokens, serving as gist tokens to store compressed content. 
Here we refer to $C$ as \textit{cache tokens} and denote $|C|$ as the \textit{cache size}.
The token $\texttt{[o]}$ is a mandatory output token, enabling continual generation based on compressed content, inspired by~\citeauthor{emnlp24_onegen}.
Finally, the enhanced data is 
\begin{align*}
\hat{Y}=\{S_1, \underline{\texttt{<w>}, C, \texttt{[o]}}, S_2, \underline{\texttt{<w>}, C, \texttt{[o]}}, \dots, S_k\},
\end{align*}
and the enhanced dataset is defined as $\hat{\mathcal{D}}=\{(X,\hat{Y})_i\}_{i=1}^{|\hat{\mathcal{D}}|}$.
For simplicity, we assume $\texttt{<w>}\in S_i$, so we omit it. 
Additionally, we use superscripts to distinguish different special tokens at different positions, such as $C^{(1)}$ and $\texttt{[o]}^{(1)}$ for tokens following $S_1$, though they are the same across different positions.

\paragraph{Thought-based Attention Mask Construction.}
To enable LLMs to learn how to compress and how to generate based on the compressed content (i.e., how to understand the compressed content), we manipulate \textit{Thought-based Mask Construction} as shown in Figure~\ref{fig:method}(b).
Specifically, let $S_{<i}=\{S_1,\dots,S_{i-1}\}$ denotes the sequence before the $i$-th thought $S_i$.

During compression, $C^{(i)}$ tokens can only attend to the question $X$, previous compressed content $\{{C,\texttt{[o]}\}}^{(<i)}$, and the current thought $S_i$, that is, 
\begin{align*}
C^{(i)} &\leftarrow \texttt{Cmp}(X, \\
&\quad \{C^{(1)}, \texttt{[o]}^{(1)}, \dots, C^{(i-1)}, \texttt{[o]}^{(i-1)}\}, S_i),
\end{align*}
where $\texttt{Cmp()}$ is compression operation.
This allows the LLM to compress the key content of $S_i$ into $C^{(i)}$.
A detailed mathematical description of $\texttt{Cmp()}$ is in Appx.~\ref{sec:app:compress}.

During generation, token $\texttt{[o]}^{(i)}$ can only attend to the question $X$ and the previous compressed content $\{C, \texttt{[o]}\}^{(\le i)}$, that is, 
\begin{align*}
S_{i+1}\leftarrow \texttt{Gen}(X,\{C^{(1)}, \texttt{[o]}^{(1)},\dots,C^{(i)},\texttt{[o]}^{(i)}\}),
\end{align*}
where $\texttt{Gen()}$ is generation operation.
This enables the LLM to continue reasoning based on the question and previous compressed content.

\paragraph{Training and Inference.}
Training objective is to maximize the following probability distribution:
\begin{align*}
&P_\theta(S_1|X) \cdot P_\theta (S_2|X,C^{(1)},\texttt{[o]}^{(1)})\cdot \dots \\
&\quad\cdot P_\theta ( S_k|X, \{ C^{(i)},\texttt{[o]}^{(i)} \}_{i=1}^{k-1}  ),
\end{align*}
where $\theta$ represents the LLM parameters.
Notably, during training, LLM is not allowed to predict the input $X$ and the special tokens $C$ and $\texttt{[o]}$.
The training samples are drawn from the $\mathcal{\hat{D}}$, and we employ an attention mask to encourage the LLM to learn to compress and comprehend the compressed content. 
The entire training process remains based on next token prediction. 
The detailed inference procedure is illustrated in Fig.~\ref{fig:intro}(b) and Fig.~\ref{fig:method}(c). 
\section{Experiments}

\subsection{Experimental Settings}

\begin{table*}[!t]
\centering
\small
\scalebox{0.7}{
\begin{tabular}{l *{4}{p{0.60cm}} *{1}{|p{0.60cm}} *{3}{p{0.60cm}} *{1}{|p{0.60cm}} *{3}{p{0.60cm}} *{1}{|p{0.60cm}} *{3}{p{0.60cm}} *{1}{|p{0.60cm}} *{3}{p{0.60cm}}}
\toprule
\multirow{4}{*}{\textbf{Method}} & \multicolumn{4}{c}{\textbf{GSM8K}} & \multicolumn{4}{c}{\textbf{MMLU}} & \multicolumn{4}{c}{\textbf{GPQA}} & \multicolumn{4}{c}{\textbf{BBH}} & \multicolumn{4}{c}{\textbf{AVG.}} \\ \cmidrule(lr){2-5} \cmidrule(lr){6-9} \cmidrule(lr){10-13} \cmidrule(lr){14-17} \cmidrule(lr){18-21}
 & Acc~$\uparrow$ & Time~$\downarrow$ & Peak~$\downarrow$ & Dep~$\downarrow$ & Acc~$\uparrow$ & Time~$\downarrow$ & Peak~$\downarrow$ & Dep~$\downarrow$ & Acc~$\uparrow$ & Time~$\downarrow$ & Peak~$\downarrow$ & Dep~$\downarrow$ & Acc~$\uparrow$ & Time~$\downarrow$ & Peak~$\downarrow$ & Dep~$\downarrow$ & Acc~$\uparrow$ & Time~$\downarrow$ & Peak~$\downarrow$ & Dep~$\downarrow$ \\ \cmidrule{1-21}
\rowcolor{mygray} \multicolumn{21}{c}{\textit{Qwen2.5-7B Series}} \\ \cmidrule{1-21}
CoT & {86.12} & {1.66} & {513} & {0.1M} & {66.50} & {1.77} & {649} & {0.2M} & {26.76} & {0.60} & {968} & {0.5M} & {65.45} & {0.68} & {570} & {0.1M} & {61.21} & {1.18} & {675} & {0.2M} \\
Distill-R1 & {81.88} & {5.60} & {844} & {1.1M} & {51.70} & {14.31} & {2483} & {7.5M} & {24.75} & {8.01} & {6718} & {31M} & {57.78} & {5.53} & {1967} & {6.0M} & {54.03} & {8.36} & {3003} & {11.3M} \\ \cmidrule{1-21}
Vanilla & {90.90} & {11.83} & {2086} & {3.9M} & {59.98} & {20.61} & {3417} & {10M} & {30.81} & {10.76} & {8055} & {39M} & {69.90} & {11.50} & {3786} & {13M} & {62.90} & {13.68} & {4336} & {16.6M} \\ 
\rowcolor{myblue}
~~+~H2O & \underline{89.92} & {22.19} & \textbf{640} & \underline{1.2M} & \underline{59.69} & {29.02} & {1024} & {3.2M} & {24.75} & {15.61} & \underline{1200} & \underline{9.8M} & \underline{70.10} & {15.61} & \textbf{1024} & \underline{3.5M} & \underline{61.12} & {20.61} & \textbf{972} & \underline{4.4M} \\
\rowcolor{myblue}
~~+~SepLLM & {30.40} & {53.52} & {1024} & {6.9M} & {10.81} & {53.45} & {1024} & {9.0M} & {0.00} & {11.65} & \textbf{1024} & {10M} & {8.08} & {26.64} & {1024} & {9.4M} & {12.32} & {36.32} & \underline{1024} & {8.9M} \\
\rowcolor{myblue}
AnLLM & {78.39} & {15.26} & {789} & {1.6M} & {54.63} & {14.13} & \underline{875} & \underline{2.0M} & {19.70} & {9.14} & {3401} & {11M} & {54.95} & {10.04} & {1303} & {3.8M} & {51.92} & {12.14} & {1592} & {4.6M} \\ \cmidrule{1-21}
\rowcolor{myblue} 
Ours~(\textit{tho.}) & \textbf{90.14} & \textbf{11.46} & \underline{676} & \textbf{1.0M} & \textbf{60.47} & \textbf{13.09} & {944} & \textbf{1.9M} & \textbf{30.30} & \underline{8.41} & {2385} & \textbf{9.3M} & \textbf{70.30} & \textbf{7.71} & \underline{1151} & \textbf{2.7M} & \textbf{62.80} & \textbf{10.17} & {1289} & \textbf{3.7M} \\ 
\rowcolor{myblue} 
Ours~(\textit{token}) & {87.11} & \underline{11.48} & {1038} & {1.5M} & {57.35} & \underline{13.80} & \textbf{489} & {3.5M} & \underline{28.28} & \textbf{8.26} & {3940} & {18M} & {62.83} & \underline{8.95} & {1884} & {5.6M} & {58.89} & \underline{10.62} & {1838} & {7.2M} \\ \cmidrule{1-21}
\rowcolor{mygray} \multicolumn{21}{c}{\textit{Llama3.1-8B Series}} \\ \cmidrule{1-21}
CoT & {85.14} & {2.15} & {550} & {0.2M} & {65.82} & {2.39} & {736} & {0.3M} & {24.75} & {0.96} & {1231} & {0.9M} & {66.46} & {0.93} & {642} & {0.2M} & {60.54} & {1.61} & {790} & {0.4M} \\
Distill-R1 & {73.62} & {2.58} & {395} & {0.1M} & {53.46} & {2.97} & {582} & {0.8M} & {20.20} & {5.24} & {3972} & {16M} & {61.21} & {0.83} & {380} & {0.2M} & {52.12} & {2.91} & {1332} & {4.4M} \\ \cmidrule{1-21}
Vanilla & {91.43} & {12.06} & {1986} & {3.0M} & {69.62} & {14.82} & {2883} & {6.9M} & {40.91} & {7.98} & {6622} & {26M} & {83.03} & {6.80} & {2793} & {5.9M} & {71.25} & {10.42} & {3571} & {10.5M} \\
\rowcolor{myblue} 
~~+~H2O & \textbf{90.45} & {20.23} & {640} & \underline{1.0M} & \textbf{65.92} & {27.11} & \underline{736} & {1.8M} & {31.81} & {12.55} & {1536} & {7.9M} & \underline{78.99} & {11.43} & {1024} & {2.1M} & \underline{66.79} & {17.83} & \underline{984} & {3.2M} \\
\rowcolor{myblue} 
~~+~SepLLM & {26.25} & {50.05} & {1024} & {5.8M} & {25.12} & {50.11} & {1024} & {7.5M} & {2.53} & {12.62} & \underline{1024} & {10M} & {14.55} & {27.14} & {1024} & {8.5M} & {17.11} & {34.98} & {1024} & {8.0M} \\
\rowcolor{myblue} 
AnLLM & {77.33} & {17.92} & \textbf{589} & {1.1M} & {58.62} & {16.53} & \textbf{589} & \textbf{1.2M} & {31.31} & {7.19} & \textbf{838} & \textbf{3.7M} & {68.89} & {9.79} & \textbf{621} & \textbf{1.6M} & {59.04} & {12.86} & \textbf{659} & \textbf{1.9M} \\ \cmidrule{1-21}
\rowcolor{myblue} 
Ours~(\textit{tho.}) & \underline{88.25} & \textbf{12.65} & \underline{629} & \textbf{0.9M} & \underline{63.39} & \textbf{14.88} & {882} & \underline{1.8M} & \textbf{36.36} & \textbf{6.38} & {1796} & \underline{6.4M} & \textbf{79.39} & \underline{7.46} & \underline{911} & \underline{1.9M} & \textbf{66.85} & \textbf{10.34} & {1055} & \underline{2.7M} \\
\rowcolor{myblue} 
Ours~(\textit{token}) & {85.52} & \underline{13.87} & {1104} & {1.7M} & {61.05} & \underline{15.85} & {1538} & {3.3M} & \underline{31.82} & \underline{6.94} & {3150} & {12M} & {74.14} & \textbf{7.43} & {1512} & {2.9M} & {63.13} & \underline{11.02} & {1826} & {4.8M} \\ \bottomrule
\end{tabular}
}

\caption{
Main results. 
The CoT is based on the instruction model, while Vanilla, AnLLM, and \ours~are based on Distill-R1. 
The light blue background indicates acceleration methods, with bold representing the best and underline the second best among them.
The Acc of Vanilla serves as the upper bound for Acc of acceleration methods. 
Dep is measured in million, Time in hours, and Peak in counts.
The compression ratio can be roughly estimated by the ratio of Dep between acceleration methods and Vanilla.
See Appendix~\ref{sec:app:metric} for more details.
Note that the results here are based on the same batch size. 
The results under the same memory budget are shown in Table~\ref{table:exp:efficiency:same_memory}.
}
\label{table:exp_main}
\end{table*}

\textbf{Baselines.}
We conduct experiments on two LLMs: Qwen2.5-7B~\citep{arxiv24_qwen2_5} and Llama3.1-8B~\citep{arxiv24_llama_3}.
To establish an upper bound performance, we perform full parameter instruction tuning using the Bespoke-Stratos-17k dataset (abbr. BS17K, with a data sample shown in Fig.~\ref{prompt:case:train}), and the fine-tuned model is denoted as \textit{Vanilla}.  
Notably, we initialize the training with the R1-Distill~\citep{arxiv25_deepseek_r1} (e.g., \texttt{DeepSeek-R1-Distill-Qwen-7B}) model, as we found that finetuning on instruction models (e.g., \texttt{Qwen2.5-7B-instruct}) yields limited improvements.
We introduce five baselines for comparison: 
two training-free acceleration methods applied to Vanilla (H2O~\citep{nips23_h2o} and SepLLM~\citep{arxiv24_sepllm}, which retain important KV Cache through specific strategies), 
one training-based method (AnLLM~\citep{acl24_anllm}), 
and two CoT~\citep{nips22_cot} baselines (prompt the instruction model and the R1-Distill model). 
More details about baselines can be found in Appx.~\ref{sec:app:exp:baseline_details}.  

\paragraph{Evaluation Metrics and Datasets.}
We evaluate \ours~on four datasets: GSM8K~\citep{arxiv21_gsm8k}, MMLU~\citep{iclr21_mmlu}, GPQA~\citep{colm24_gpqa}, and BBH~\citep{acl23_bbh}. 
For MMLU and BBH, we randomly sample a portion of the data for evaluation. 
The evaluation focuses on both \textit{effectiveness} and \textit{efficiency}. 
For effectiveness, we use accuracy as the evaluation metric (\textit{Acc}); 
for efficiency, we employ three metrics: inference time (\textit{Time}), the peak number of tokens in the context during inference (\textit{Peak}), and the sum of \textit{dependency} of each generated token on previous tokens during the generation (\textit{Dep}). 
Fig.~\ref{fig:exp:metric} visualizes the Peak and Dep metrics, where the value of Dep equals the area enclosed by the lines. 
The Dep metric characterizes the amount of information used during inference, with smaller values indicating more significant compression. 
We aim to compare the other three metrics under similar Dep values. 
It is important to note that Peak characterizes a momentary state, while Dep characterizes the entire inference process, so there is no direct correlation between the two. 
For more details about Dep, please refer to Appx.~\ref{sec:app:metric}.  

\paragraph{Implementation.}
For~\ours, we design two different segmentation functions $\texttt{Seg()}$. 
At the token level, we compress every 6 tokens into 2 tokens, i.e., $|C|=2$, denoted as ``ours (\textit{token})''. 
At the thought level, we use ``{\textbackslash n\textbackslash n}'' as a delimiter to simply segment the B17K data into several thoughts, denoted as ``ours (\textit{tho.})''. 
For the Qwen, we compress a thought into 9 tokens, i.e., $|C|=9$; 
for the Llama, we compress a thought into 7 tokens, i.e., $|C|=7$. 
In all experiments, we use greedy decoding with a maximum output length of 10240 tokens.
Please refer to Appx.~\ref{sec:app:exp} for more details.

\subsection{Main Results}
In Tab.~\ref{table:exp_main}, we report the results of four evaluation metrics for two models on four datasets.
\textbf{Key observations include:}
\textbf{\textit{1)}} 
Distill-R1 performs worse than CoT across all datasets because its weaker instruction-following ability~\citep{arxiv25_think_fail} prevents effective answer extraction using rules, even with attempts at evaluation using an LLM evaluator. 
However, this issue is not the focus of this paper.
\textbf{\textit{2)}} H2O effectively reduces memory usage while maintaining the performance of the vanilla, indicating that the greedy eviction policy is effective in long-text generation tasks. 
However, H2O significantly increases inference time compared to Vanilla, with an average increase of 51\% ($(20.61-13.68)/13.68\approx 0.51$) on Qwen and 72\% on Llama. 
This is attributed to the token-wise eviction policy of H2O, which introduces additional overhead for each generated token.
\textbf{\textit{3)}} SepLLM performs the worst in terms of performance, gradually losing language ability during generation, which results in the inability to output termination tokens and thus leads to excessive inference time.
\textbf{\textit{4)}} Compared to H2O, \ours~(\textit{tho.}) achieves similar performance with lower Dep values (i.e., similar compression rate), while reducing inference time by an average of 52\% on Qwen and 41\% on Llama. 
Additionally, \ours~(\textit{tho.}) retains higher accuracy and faster inference speed compared to AnLLM.

Based on these observations, \textbf{we draw the following conclusions:}
\textbf{\textit{1)}} {B17K is an effective dataset.}
We find Vanilla outperforms CoT and Distill-R1 on most datasets, indicating that B17K is an effective dataset that mitigates the repetition issue in Distill-R1 through SFT.
\textbf{\textit{2)}} {\ours~is effective and achieves a good balance between effectiveness and efficiency in inference.}
Specifically, on the Qwen, \ours~sacrifices 1\% accuracy but saves 26\% time, reduces the peak tokens by 70\%, and decreases Dep. by 78\% (i.e., achieves a 16.6/3.7=4.5x compression ratio). 
On the Llama, it sacrifices 6\% accuracy but saves 1\% inference time, reduces the peak tokens by 70\%, and decreases Dep. by 74\% (i.e., achieves a 10.5/2.7=3.9x compression ratio).
\textbf{\textit{3)}} {The segmentation function is vital for~\ours.} 
The thought-level segmentation function outperforms the token-level, with accuracy improvements of 6.2\% on Qwen and 5.6\% on Llama. 
This suggests that token-level segmentation leads to the loss of semantic boundaries.

\begin{figure*}[!th] 
    \centering
    \scalebox{0.9}{
    \includegraphics[width=1\linewidth]{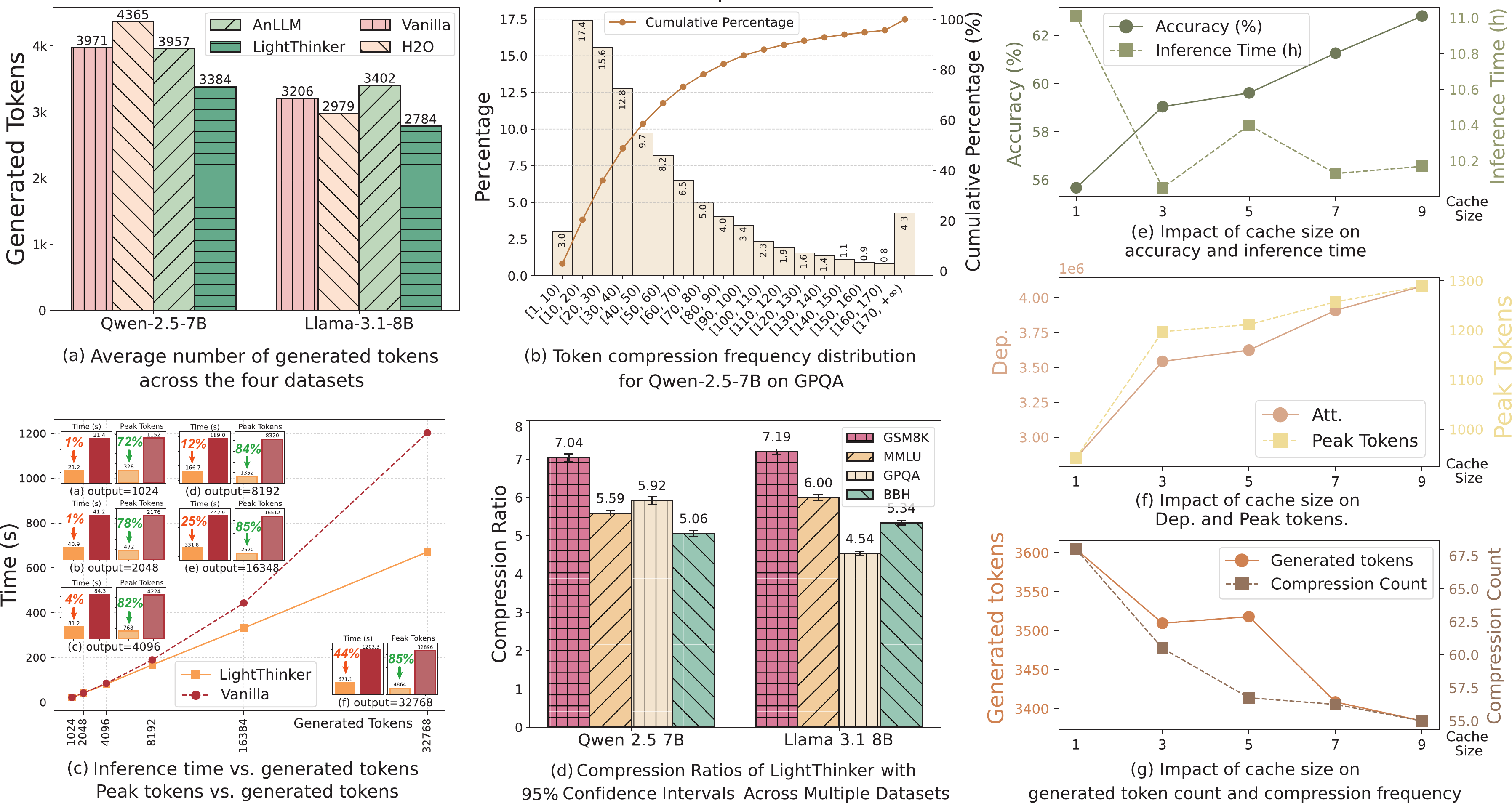} 
    }
    \caption{
    Efficiency Analysis and Ablation Results.
    Fig.(a) represents the average tokens generated by the respective model on the specified dataset. 
    Fig.(b) shows the percentage of tokens falling within specified ranges, while the cumulative percentage curve illustrates the total proportion of tokens up to each range. 
    Fig.(c) illustrates the relationship between the number of generated tokens and inference time. 
    Each subplot displays the inference time and peak token for various numbers of output tokens. 
    Fig.(d) represents the average compression ratios with 95\% confidence intervals indicated by error bars. 
    Fig.(e-f) examines the impact of cache size (i.e., $|C|$) on accuracy, Dep, inference time, peak tokens, generated tokens, and compression frequency.
    }
    \label{fig:exp:efficient}
    \vspace{-3mm}
\end{figure*}

\vspace{-1mm}
\subsection{Efficiency}
For clarity, ``\ours'' hereafter denotes \ours~(\textit{tho.}).
In this section, we conduct an in-depth analysis of \ours's efficiency, focusing on the following four questions:
\begin{table}[]
\centering
\small
\scalebox{0.85}{
\begin{tabular}{cccccc}
\toprule
             & GSM8K & MMLU  & GPQA  & BBH   & AVG   \\ \midrule
Vanilla      & 11.83 & 20.61 & 10.76 & 11.50 & 13.68 \\
LightThinker & \textbf{6.73}  & \textbf{7.44}  & \textbf{3.86}  & \textbf{3.97}  & \textbf{5.50}  \\ \bottomrule
\end{tabular}
}
\caption{
 Inference time comparison (in hours) for Vanilla and \ours~on the Qwen model across four datasets under the same memory budget.
}
\vspace{-0.5cm}
\label{table:exp:efficiency:same_memory}
\end{table}

\paragraph{How does \ours~accelerate under same memory budget?}
Inference efficiency is measured through memory consumption and inference speed.
Tab.~\ref{table:exp_main} highlights \ours’s ability to significantly reduce memory consumption at the same batch size. 
In practice, this reduction allows for larger batch sizes under the same memory budget, improving throughput. 
Experiments on four datasets with the Qwen model, conducted under the same memory constraints, show that \ours~reduces inference time by an average of 2.5× compared to Vanilla, as demonstrated in Tab.~\ref{table:exp:efficiency:same_memory}.
This indicates that \ours~not only reduces both memory and time overhead at the same batch size (as shown in Tab.~\ref{table:exp_main}) but also significantly lowers time overhead under the same memory budget, thereby improving throughput.

\paragraph{Does \ours~generate more tokens compared to Vanilla?}
Fig.~\ref{fig:exp:efficient}(a) shows the average number of generated tokens for H2O, AnLLM, LightThinker, and Vanilla across four datasets (others in the Appx.~\ref{sec:app:exp:additional_results}). 
We observe that:
1) \ours~is the only method that reduces the number of generated tokens compared to Vanilla, with an average reduction of 15\% on Qwen and 13\% on Llama. 
This is one of the reasons for its faster inference speed.
2) H2O increases token generation by 10\% on Qwen but reduces it by 7\% on Llama. 
Despite the reduction in tokens for Llama, the inference time still increases as shown in Tab.~\ref{table:exp_main}, indicating that its eviction policy accumulates additional overhead as token generation grows.

\paragraph{What is the compression ratio of \ours?}
Fig.~\ref{fig:exp:efficient}(d) illustrates the compression ratio across four datasets, 
Tab.~\ref{table:exp:efficiency:comp_count} reports the average compression counts, 
and Fig.~\ref{fig:exp:efficient}(b) shows the distribution of compressed token counts for GPQA using Qwen (other datasets are in the Appx.~\ref{sec:app:exp:additional_results}). 
We find that:
1) Compression counts and ratios are more closely related to downstream tasks than to the model itself. 
Simple tasks like GSM8K exhibit lower compression counts and higher compression ratios, while complex tasks like GPQA require more frequent compressions and smaller compression ratios. 
2) The distribution of compressed token counts follows a long-tail pattern. 

\paragraph{How efficient is \ours~in memory usage and inference for long-text generation?}
Fig.~\ref{fig:exp:efficient}(c) shows the inference time and peak tokens of \ours~and Vanilla as a function of output token length.
We set the prompt length to 125 and compressed 56 tokens to 8 tokens (i.e., $|C|=7$).
We observe that:
1) Our method significantly reduces inference time. 
For example, when generating 32K tokens, the inference time is reduced by 44\%. 
For shorter texts (from 1K to 4K tokens), the reduction is more modest, ranging from 1\% to 4\%.
2) Even for shorter texts, \ours~substantially reduces peak tokens. 
For instance, when generating 1K tokens, peak tokens are reduced by 72\%, and for 32K tokens, it is reduced by 85\%.

\begin{table}[]
\centering
\small
\scalebox{1}{
\begin{tabular}{ccccc}
\toprule
      & GSM8K & MMLU & GPQA & BBH \\ \midrule
Qwen  & 20    & 37   & 115  & 48  \\
Llama & 26    & 47   & 139  & 55  \\ \bottomrule
\end{tabular}
}
\caption{
Statistics of the average number of compressions per dataset for \ours.
}
\label{table:exp:efficiency:comp_count}
\vspace{-2mm}
\end{table}
\subsection{Ablation}
\label{ablation}
\paragraph{Decoupled Token and Attention Mask Mode.}
\ours~differs from AnLLM in two key aspects: the decoupled token design and the attention mask as shown in Figure~\ref{fig:comp}. 
To validate the effectiveness of these mechanisms, we conduct ablation experiments.  
As shown in Table~\ref{table:exp:ablation:attention}, under the same cache size setting and using AnLLM's attention mask mechanism (``AnLLM'' vs. ``Ours ($|C|=1$, T)''), the decoupled design improves accuracy by 2\%. 
Further adopting \ours's attention mask mode yields an additional 7\% improvement. 
These results demonstrate the effectiveness of both the decoupled token and the attention mask mode in \ours.

\paragraph{Cache Size.}
We varied $|C|$ in $\{1,3,5,7,9\}$ to observe its impact on accuracy, inference time, dependency (i.e., Dep), peak tokens, generated token count, and compression frequency. 
Fig.~\ref{fig:exp:efficient}(e-g) illustrate these trends on the Qwen model. 
We observe that:  
1) As shown in Figure~\ref{fig:exp:efficient}(e), increasing the cache size significantly improves accuracy while reducing inference time. 
This indicates that a larger cache size mitigates information loss caused by compression.  
2) As shown in Figure~\ref{fig:exp:efficient}(g), increasing the cache size reduces both the compression frequency and the number of generated tokens.
3) Combining Fig.~\ref{fig:exp:efficient}(e) and Fig.~\ref{fig:exp:efficient}(g), we find that a smaller cache size leads to more frequent generation and compression to retain more information, while a larger cache size reduces this frequency.  

\begin{figure}[!t] 
    \centering
    \scalebox{1}{
    \includegraphics[width=1\linewidth]{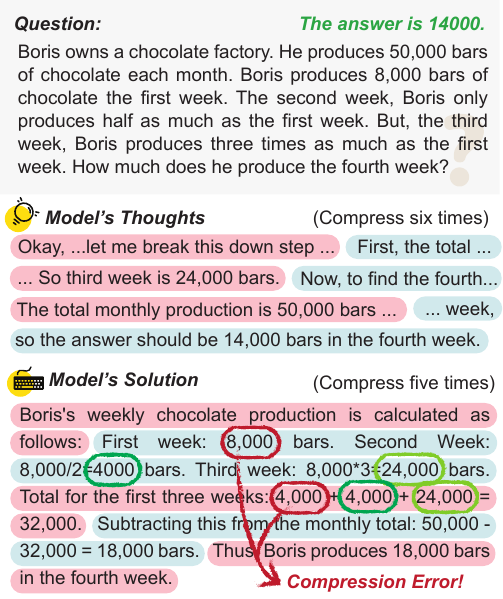} 
    }
    \caption{
    Case Study. 
    The figure illustrates partial inference results of a case from GSM8K.
    See App.~\ref{sec:app:exp:additional_results} for the complete content.  
    Pink and light blue backgrounds are used to distinguish adjacent compression processes, where each color represents one compression.}
    \label{fig:exp:case}
\end{figure} 
\subsection{Case Study}

Fig.~\ref{fig:exp:case} illustrates a failure case from the GSM8K dataset. 
We observe that although the LLM arrives at the correct answer during the thinking process (see \texttt{Model's Thoughts} field in the Fig.~\ref{fig:exp:case}), it makes an error in the final output (see \texttt{Model's Solution} field in the Figure). 
Specifically, in the third sentence of the $\texttt{Model's Solution}$ field, the first occurrence of ``4000'' is incorrect. 
This indicates that information loss occurred during the second compression step (theoretically, ``8000'', ``4000'', and ``24000'' should have been compressed, but the LLM only compressed ``4000'' and ``24000''), leading to subsequent reasoning errors. 
Such errors occur frequently in the GSM8K dataset, suggesting that the current compression method is not sufficiently sensitive to numerical values. 

\begin{table}[]
\centering
\small
\scalebox{0.8}{
\begin{tabular}{lccccc}
\toprule
 & \multicolumn{1}{l}{GSM8K} & \multicolumn{1}{l}{MMLU} & \multicolumn{1}{l}{GPQA} & \multicolumn{1}{l}{BBH} & \multicolumn{1}{l}{AVG} \\ \cmidrule{1-6}
AnLLM & \underline{78.39} & 54.63 & 19.70 & 54.95 & 51.92 \\
Ours (|C|=1, T)  & 78.32 & \underline{58.23} & \underline{20.71} & \underline{55.35} & \underline{53.15} \\
Ours (|C|=1, F) & \textbf{80.21} & \textbf{58.23} & \textbf{22.22} & \textbf{62.02} & \textbf{55.67} \\ \bottomrule
\end{tabular}
}
\caption{
Ablation results on the Qwen, reporting accuracy on four datasets. 
``T'' denotes the use of AnLLM's attention mask mechanism, while ``F'' indicates the use of \ours's attention mask mechanism.
}
\label{table:exp:ablation:attention}
\vspace{-1mm}
\end{table}

\vspace{-1mm}
\section{Related Work}
\vspace{-1mm}
Current research on accelerating the inference process of LLMs primarily focuses on three categories of methods: \textit{Quantizing Model}, \textit{Generating Fewer Tokens}, and \textit{Reducing KV Cache}. 
Quantizing Model includes both parameter quantization~\cite{mlsys24_awq} and KV Cache quantization~\citep{icml24_kivi}. 
Notably, generating long texts and understanding long-text represent distinct 
scenarios; 
therefore, acceleration methods specifically targeting the long-text generation phase (e.g., pre-filling stage acceleration techniques~\citep{emnlp23_autocompressors,iclr24_icae,emnlp23_llmlingua,iclr25_activation_beacon,nips24_snapkv,arxiv24_pyramidkv} are not discussed here. 
Due to page limits, we focus on the last one. 
See Appx.~\ref{sec:app:related_work} for other details.

\paragraph{Reducing KV Cache.}
This category can be divided into two types of strategies: pruning-based KV Cache selection in discrete space and merging-based KV Cache compression in continuous space.
1) \textit{Pruning-Based Strategies}.
Specific eviction policies~\citep{nips23_h2o, iclr24_streamingllm,arxiv24_sepllm} are designed to retain important tokens during inference.
2) \textit{Merging-Based Strategies}.
Anchor tokens are introduced, and LLMs are trained to compress historically important information into these tokens, thereby achieving KV Cache merging~\citep{acl24_anllm}.
Both strategies require intervention during inference. 
The key difference is that the first strategy is training-free but applies the eviction policy for every generated token, while the second is a training-based method and allows the LLM to decide when to apply the eviction policy.

\vspace{-1mm}
\section{Conclusion}
\vspace{-1mm}
In this paper, we present \ours, a new approach to enhance the efficiency of LLMs in complex reasoning tasks by dynamically compressing intermediate thoughts during generation. 
By training the LLM to learn when and how to compress verbose thought steps into compact representations, \ours~significantly reduces memory overhead and computational costs while maintaining competitive accuracy. 
We introduce the \textit{Dependency} (abbr., Dep) metric to quantify the degree of compression across different accelerating methods.
Extensive experiments demonstrate that \ours~is an effective approach to balancing efficiency and performance. 





\clearpage
\section*{Limitations}
Although \ours~has shown remarkable advancements in memory optimization and inference speed enhancement, certain limitations warrant careful consideration:
\begin{enumerate}




    \item The number of cache tokens is fixed during training and must remain consistent during inference. 
    The generalization capability of these token representations is uncertain. 
    For instance, whether representations trained with 3 tokens can extrapolate to scenarios requiring more tokens during inference.

    \item The design of the segmentation function is relatively simplistic, relying on rule-based methods. 
    Future work could investigate more advanced segmentation strategies.

    \item The performance of \ours~on tasks such as novel generation, code generation, and multi-turn dialogue remains unassessed.


\end{enumerate}

\section*{Acknowledgement}
We would like to express our sincere gratitude to the anonymous reviewers for their thoughtful and constructive feedback. This work was supported by the National Natural Science Foundation of China (No. 62576307, No. NSFCU23B2055, No. NSFCU19B2027), the Fundamental Research Funds for the Central Universities (226-2023- 00138), Ningbo Natural Science Foundation (2024J020), Yongjiang Talent Introduction Programme (2021A-156-G), and Information Technology Center and State Key Lab of CAD\&CG, Zhejiang University. This work was supported by Ant Group and Zhejiang University - Ant Group Joint Laboratory of Knowledge Graph.

\bibliography{custom}




\clearpage
\appendix

\section*{Appendix}
\section{Metric: \texttt{Dependency}}
\label{sec:app:metric}
\begin{figure}[!htbp] 
    \centering
    \scalebox{1}{
    \includegraphics[width=1.0\linewidth]{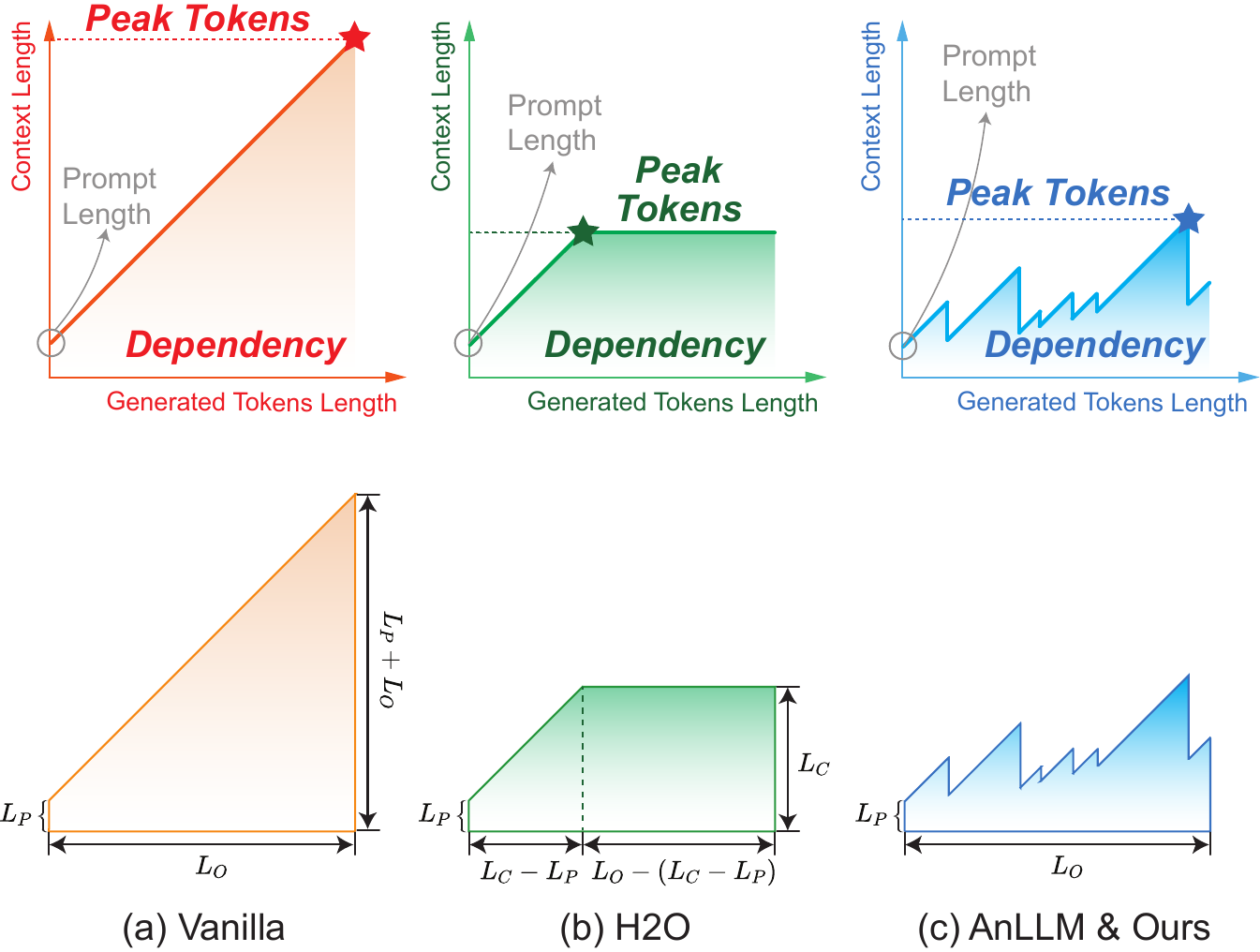} 
    }
    \caption{Illustration of the metric Dependency.}
    \label{fig:app:metric}
\end{figure} 
\subsection{Motivation}
\ours~and AnLLM~\citep{acl24_anllm} are dynamic compression methods, meaning the number of compressions and the compression ratio are determined by the LLM itself rather than being predefined hyperparameters. 
In contrast, H2O~\citep{nips23_h2o} and SepLLM~\citep{arxiv24_sepllm} allow users to set hyperparameters to control the maximum number of tokens retained during inference. 
This fundamental difference makes it challenging to directly and fairly compare dynamic compression methods like \ours~and AnLLM with KV cache compression approaches like H2O and SepLLM.  

Traditionally, KV cache compression methods are compared by setting the same maximum peak token count, but this metric becomes inadequate in our context. 
As illustrated in Figure~\ref{fig:app:metric}, which shows the relationship between generated tokens and context length for Vanilla, H2O, and \ours, 
\ours~occasionally exceeds H2O in peak token count. 
However, this metric is misleading because \ours's peak memory usage occurs only momentarily, while H2O maintains a consistently high token count over time.  

Moreover, previous KV cache compression methods often compress prompt parts only and assume a fixed prompt length, allowing compression ratios to be predefined. 
In our setting, however, the output is also needed to be compressed.
The output token count is unknown, making it impossible to preset a global compression ratio. 
Consequently, relying solely on maximum peak token count as a comparison metric is insufficient.  

To address these challenges, we propose a new metric called \textit{Dependency}, which quantifies the total amount of information dependencies during the generation process. 
This metric enables fair comparisons between dynamic compression methods and traditional KV cache compression approaches by ensuring evaluations are conducted under similar effective compression ratios.  
\subsection{Definition}
We introduce the \textbf{Dependency} (abbr., Dep) metric, defined as the sum of dependencies of each generated token on previous tokens during the generation of an output. 
Geometrically, it represents the area under the curve in Figure~\ref{fig:app:metric}. 
Dependency can be calculated either from its definition or through its geometric interpretation. 
Here, we focus on the geometric approach. 
Let the initial prompt length be \( L_P \), the model's output length be \( L_O \), and the maximum context length set by KV cache compression methods be \( L_C \).  

\textbf{Dependency for Vanilla.}
The area under Vanilla's curve forms a right trapezoid, calculated as:  
\[
\begin{aligned}
\texttt{Dependency} &= \frac{(L_P + L_P + L_O) \times L_O}{2} \\
&= \frac{{L_O}^2}{2} + L_P \times L_O
\end{aligned}
\]

\textbf{Dependency for H2O.}
The area under H2O's curve consists of a trapezoid (left part in Figure~\ref{fig:app:metric}(b)) and a rectangle (right part in Figure~\ref{fig:app:metric}(b)):  
\[
\begin{aligned}
S_\texttt{Trapezoid} &= \frac{(L_P + L_C) \times (L_C - L_P)}{2} \\
S_\texttt{rectangle} &= L_C \times (L_O - L_C + L_P) \\
\texttt{Dependency} &= S_\texttt{Trapezoid} + S_\texttt{rectangle} \\
&= \frac{2L_PL_C + 2L_OL_C - {L_P}^2 - {L_C}^2}{2}
\end{aligned}
\]

\textbf{Dependency for \ours~and AnLLM.}
For \ours~and AnLLM, Dependency does not have a closed-form solution and must be computed iteratively based on its definition. 

\subsection{Application}
\textbf{Value of Dependency.}
A higher Dependency value indicates that more tokens need to be considered during generation, reflecting greater information usage.
Conversely, a lower Dependency value suggests a higher effective compression ratio.  

\textbf{Dependency Ratio.}
By dividing the Dependency of an accelerated method by that of Vanilla, we obtain the compression ratio relative to Vanilla. For example, in Table~\ref{table:exp_main}'s ``Avg.'' column, Vanilla's Dependency is 16.6M, H2O's is 4.4M, and \ours's is 3.7M. 
Thus, H2O achieves a compression ratio of \( \frac{16.6}{4.4} \approx 3.8 \), while \ours~achieves \( \frac{16.6}{3.7} \approx 4.5 \).  

This metric provides a unified framework for evaluating both dynamic and static compression methods, ensuring fair and meaningful comparisons.


\begin{figure}[!htbp] 
    \centering
    \scalebox{1}{
    \includegraphics[width=1.0\linewidth]{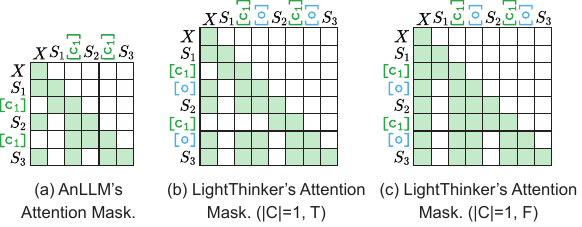} 
    }
    \caption{Illustration of Attention Mask in Table~\ref{table:exp:ablation:attention}.}
    \label{fig:app:attention_mode}
\end{figure} 
\section{Mathematical Description of Compression}
\label{sec:app:compress}
In this section, we provide a detailed formulation of the compression operation introduced in Section~\ref{sec:method:method}.

\paragraph{Notation.} 
During compression, the context can be divided into three segments: 
1. The sequence that remains in the context without being compressed, denoted as $Pre := \{X, \{C^{(1)}, \texttt{[o]}^{(1)}\dots,C^{(i-1)},\texttt{[o]}^{i-1}\}\}$, with the number of tokens represented by $N$; 
2. The thought sequence to be compressed, defined as $Tho:=S_i$, with the number of tokens denoted by $T$; 
3. The sequence storing the compressed content, $C:=C^{(i)}$, with its length represented by $|C|$.

\paragraph{Compression Operation.} 
Here, we describe the compression operation at a specific layer, focusing on the information passed to the sequence $C$. 
According to the definition of self-attention~\citep{nips17_transformer}, the attention matrix for the sequence $C$ with respect to other content is calculated as:
\[
A=\text{Softmax}(\text{mask}(\frac{Q^{C}[K^{Pre}:K^{Tho}:K^{C}]^{\top}}{\sqrt{d}}))
\]
where $[:]$ denotes the concatenation operation, $\text{mask}(\cdot)$ represents the attention mask corresponding to the ``Thought-based Attention Mask Construction'' in Section~\ref{sec:method:method}, 
$K^{Pre}, V^{Pre}\in \mathbb{R}^{N\times d}$, 
$K^{Tho}, V^{Tho}\in \mathbb{R}^{T\times d}$, 
$K^{C},V^{C}\in \mathbb{R}^{|C|\times d}$, $Q^{C}\in \mathbb{R}^{|C|\times d}$, 
and $d$ is the hidden dimension. The matrix $A\in \mathbb{R}^{|C|\times (N+T+|C|)}$ describes the attention of sequence $C$ to other content. 
The \textit{values} of the other sequences are then weighted and summed according to the attention matrix:
\[
H=A\times [V^{Pre}:V^{Tho}:{V^C}]
\]
where $[V^{Pre}:V^{Tho}:V^{C}]\in \mathbb{R}^{(N+T+|C|)\times d}$,
and thus $H\in \mathbb{R}^{|C|\times d}$. 
At this point, the information from the current $Tho$ is preserved in $H$. 
Through training, the model learns to selectively retain useful information from $Tho$ in $H$. 
$H$ is then stored in the KV Cache after passing through an MLP and the next layer's projection.

\section{Experiment}
\label{sec:app:exp}
\subsection{Training Data}
\label{sec:app:exp:train_data_case}
Examples of training samples are shown in Figure~\ref{prompt:case:train}.

\subsection{Baseline Details}
\label{sec:app:exp:baseline_details}
\textbf{H2O}~\citep{nips23_h2o} is a training-free acceleration method that greedily retains tokens with the highest cumulative attention values from historical tokens. 
It includes two hyper-parameters: the maximum number of tokens and the current window size (i.e., \texttt{local\_size}). 
The maximum number of tokens for each task is listed in the ``Peak'' column of Table~\ref{table:exp_main}, and the \texttt{local\_size} is set to half of the maximum number of tokens. 
The experimental code is implemented based on \url{https://github.com/meta-llama/llama-cookbook}.

\textbf{SepLLM}~\citep{arxiv24_sepllm} is another training-free acceleration method that considers tokens at punctuation positions as more important. 
It includes four parameters: the maximum number of tokens is set to 1024, \texttt{local\_size} is set to 256, \texttt{sep\_cache\_size} is set to 64, and \texttt{init\_cache\_size} is set to 384. 
We also tried another set of parameters (\texttt{init\_cache\_size}=4, \texttt{sep\_cache\_size}=64, \texttt{local\_size}=720, maximum number of tokens=1024), but found that the first set of parameters performed slightly better.

\textbf{AnLLM}~\citep{acl24_anllm} is a training-based method that shares a similar overall approach with \ours~but accelerates by saving historical content in anchor tokens. 
The specific differences between the two are detailed in Section~\ref{sec:app:discussion}.

\subsection{Training Details}
\label{sec:app:exp:training_details}
Both \textbf{Vanilla} and \textbf{AnLLM} are trained on the B17K~\citep{bespoke_stratos_train_dataset} dataset using the R1-Distill~\citep{arxiv25_deepseek_r1} model for 5 epochs, while \ours~is trained for 6 epochs. 
The maximum length is set to 4096, and a cosine warmup strategy is adopted with a \texttt{warmup\_ratio} of 0.05. 
Experiments are conducted on 4 A800 GPUs with DeepSpeed ZeRo3 offload enabled. 
The batch size per GPU is set to 5, and the gradient accumulation step is set to 4, resulting in a global batch size of 80. 
The learning rate for Vanilla is set to 1e-5, while for AnLLM and \ours, it is set to 2e-5.

\subsection{Evaluation Details}
\label{sec:app:exp:evaluation_details}
For the CoT in Table~\ref{table:exp_main}, the prompts used are shown in Figure~\ref{prompt:system:base} and Figure~\ref{prompt:task:base}. 
For the R1-Distill model, no system prompt is used, and the task-specific prompts are shown in Figure~\ref{prompt:task:r1}. 
Vanilla, H2O, SepLLM, AnLLM, and \ours~share the same set of prompts, with the system prompt shown in Figure~\ref{prompt:system:vanilla} and downstream task prompts shown in Figure~\ref{prompt:task:r1}. 
The options for MMLU~\citep{iclr21_mmlu} and GPQA~\citep{colm24_gpqa} multiple-choice questions are randomized.

\subsection{Additional Results}
\label{sec:app:exp:additional_results}
\begin{figure*}[!htbp] 
    \centering
    \scalebox{0.8}{
    \includegraphics[width=1.0\linewidth]{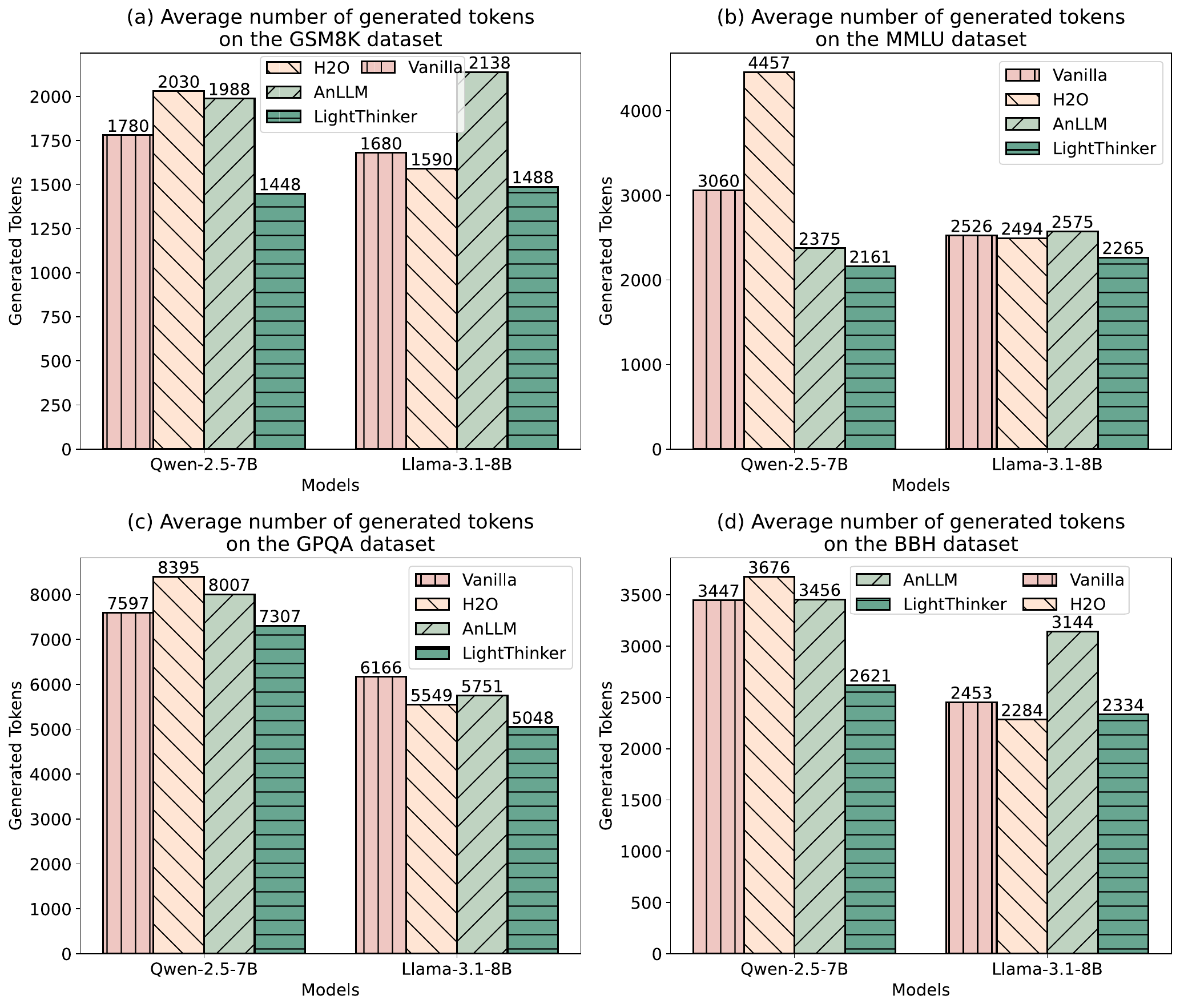} 
    }
    \caption{Average number of generated tokens.}
    \label{fig:app:token}
\end{figure*}

\begin{figure*}[!htbp] 
    \centering
    \scalebox{1}{
    \includegraphics[width=0.76\linewidth]{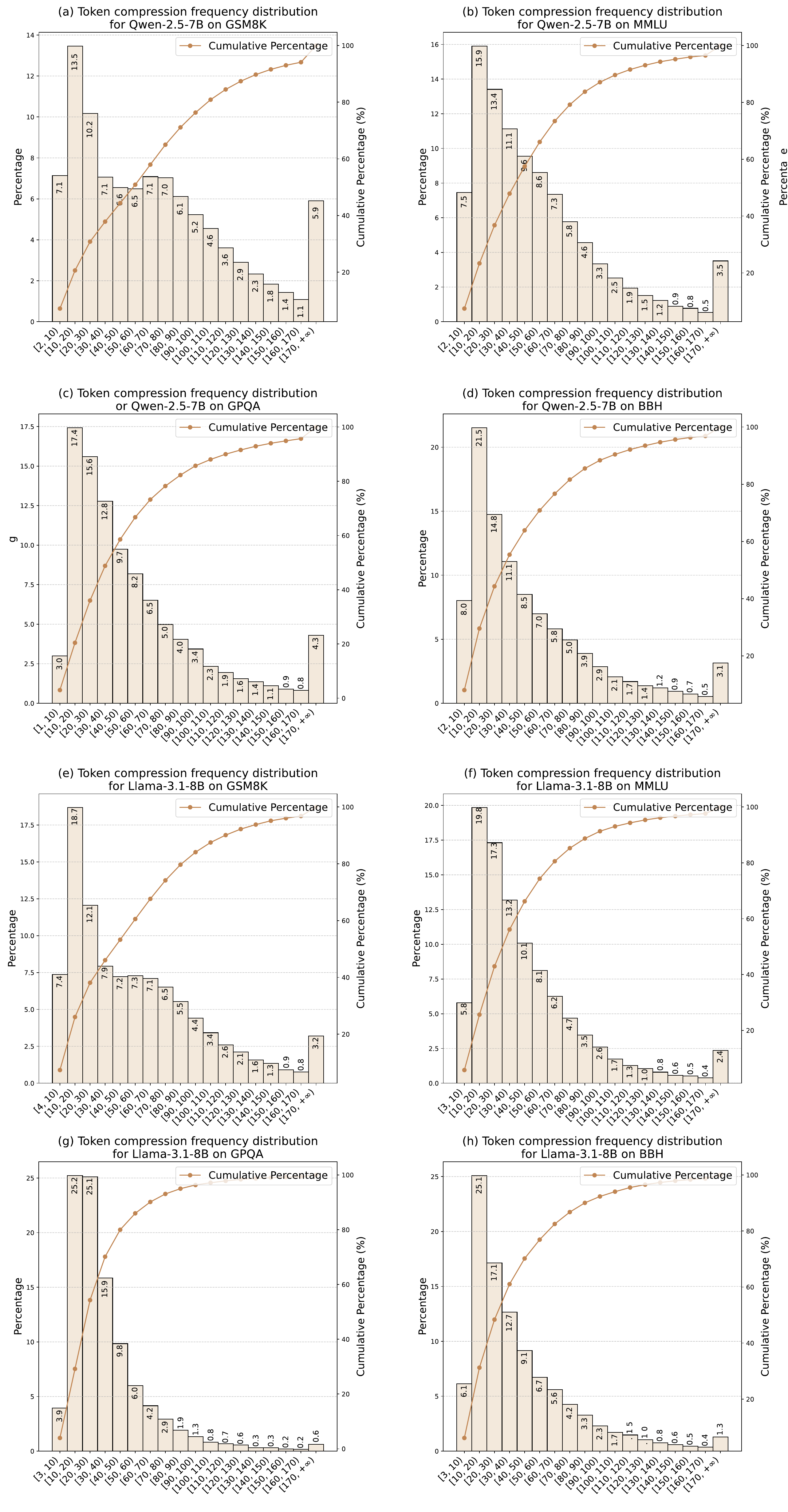} 
    }
    \caption{Token compression frequency distribution for \ours.}
    \label{fig:app:frequency}
\end{figure*}

Figure~\ref{fig:app:token} compares the number of tokens generated by two models across different datasets. 
Figure~\ref{fig:app:frequency} shows the distribution of compressed lengths for LightThinker on two models and four datasets. Figure~\ref{fig:app:attention_mode} illustrates the attention masks for the baselines in Table~\ref{table:exp:ablation:attention}.
Figure~\ref{prompt:case} shows a complete case in Figure~\ref{fig:exp:case}.

\section{Related Work}
\label{sec:app:related_work}

Current research on accelerating the inference process of large language models (LLMs) primarily focuses on three categories of methods: \textit{Quantizing Model}, \textit{Generating Fewer Tokens}, and \textit{Reducing KV Cache}. 
Quantizing Model includes both parameter quantization~\cite{mlsys24_awq,nips22_8bit} and KV Cache quantization~\citep{icml24_kivi,nips24_kvquant}, while this section will concentrate on the latter two categories. 
It is important to note that generating long texts and understanding long texts represent distinct application scenarios; 
therefore, acceleration methods specifically targeting the long-text generation phase (e.g., pre-filling stage acceleration techniques such as AutoCompressor~\citep{emnlp23_autocompressors}, ICAE~\citep{iclr24_icae}, LLMLingua~\citep{emnlp23_llmlingua}, Activation Beacon~\citep{iclr25_activation_beacon}, SnapKV~\citep{nips24_snapkv}, and PyramidKV~\citep{arxiv24_pyramidkv}) are not discussed here. 
Below is a detailed overview of the last two categories.

\textbf{Generating Fewer Tokens.}
This category can be further divided into three strategies based on the number and type of tokens used during inference.
1) \textit{Discrete Token Reduction}. 
Techniques such as prompt engineering~\citep{arxiv24_tale,arxiv24_break_the_chain,arxiv24_concise_thoughts}, instruction fine-tuning~\citep{nips24_skip_steps,arxiv24_c3ot}, or reinforcement learning~\citep{arxiv25_related_work_rl1,arxiv25_o1_pruner} are used to guide LLMs to use fewer discrete tokens during inference.
For example, TALE~\citep{arxiv24_tale} prompts LLMs to complete tasks under a predefined token budget. 
\citeauthor{arxiv25_related_work_rl1} construct specific datasets and employ reinforcement learning reward mechanisms to encourage models to generate concise and accurate outputs, thereby reducing token usage.
2) \textit{Continuous Token Replacement}.
These methods~\citep{arxiv24_coconut,arxiv24_ccot} explore using continuous-space tokens instead of traditional discrete vocabulary tokens. 
A representative example is CoConut~\citep{arxiv24_coconut}, which leverages Curriculum Learning to train LLMs to perform inference with continuous tokens.
3)\textit{No Token Usage}.
By internalizing the inference process between model layers, the final answer is generated directly during inference without intermediate tokens~\citep{arxiv24_icot,arxiv23_kd_cot}.
These three strategies are implemented after model training and do not require additional intervention during inference. 
Technically, the acceleration effect of these methods increases sequentially, but at the cost of a gradual decline in the generalization performance of LLMs. 
Additionally, the first strategy does not significantly reduce GPU memory usage.

\textbf{Reducing KV Cache.}
This category can be divided into two types of strategies: pruning-based KV Cache selection in discrete space and merging-based KV Cache compression in continuous space.
1) \textit{Pruning-Based Strategies}.
Specific eviction policies~\citep{nips23_h2o, iclr24_streamingllm,arxiv24_sepllm,arxiv24_scope} are designed to retain important tokens during inference.
For example, StreamingLLM~\citep{iclr24_streamingllm} considers the initial sink tokens and the most recent tokens as important.
H2O~\citep{nips23_h2o} focuses on tokens with high historical attention scores.
SepLLM~\citep{arxiv24_sepllm} emphasizes tokens corresponding to punctuation marks.
2) \textit{Merging-Based Strategies}.
Anchor tokens are introduced, and LLMs are trained to compress historically important information into these tokens, thereby achieving KV Cache merging~\citep{acl24_anllm}.
Both strategies require intervention during inference. 
The key difference is that the first strategy is training-free but applies the eviction policy for every generated token, while the second strategy is a training-based method and allows the LLM to decide when to apply the eviction policy.

\section{Discussions}
\label{sec:app:discussion}
\subsection{Difference between LightThinker and AnLLM}
AnLLM~\citep{acl24_anllm} is a work from 2023, at which time the concept of long-cot~\citep{arixv24_o1,arxiv25_deepseek_r1} did not exist. 
AnLLM itself focuses more on prompt compression rather than output compression. 
Additionally, our method decouples compression and generation, allowing for scaling the number of cache tokens—something AnLLM cannot do. 
Therefore, our work is only related to AnLLM in that both use sparse attention~\citep{nips23_h2o,nips24_snapkv} to speed up processes, but they are not similar works.

\begin{figure}[!htbp] 
    \centering
    \scalebox{1}{
    \includegraphics[width=1.0\linewidth]{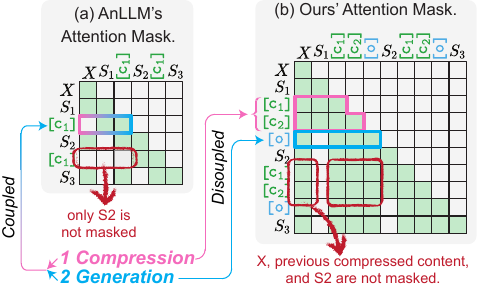} 
    }
    \caption{
    Contrast of AnLLM and ours.
    Two differences are marked: one with a red box, and the other with blue and pink arrows.
    }
    \label{fig:comp}
\end{figure}

AnLLM is a method related to ours.
In Figure~\ref{fig:comp}, we compare the differences in Attention Mask between \ours~and AnLLM:
1) \textit{Decoupling Generation and Compression.}
In AnLLM, the $\texttt{[c}_\texttt{i}\texttt{]}$ token is tasked with both compressing historical information and generating subsequent content, as shown by the blue and pink arrows in Fig.~\ref{fig:comp}. 
This design tightly couples generation and compression. 
In contrast, \ours~decouples these tasks: the $\texttt{[c}_\texttt{i}\texttt{]}$ token solely compresses historical information, while the $\texttt{[o]}$ token performs reasoning based on the compressed content.
2) \textit{Context Visibility during Compression.}
AnLLM can only access the current thought during compression.
\ours, however, allows access to $X$, historical compressed content, and the current thought during compression, thereby enhancing contextual understanding.
Ablation experiments in Section~\ref{ablation} demonstrate that these designs significantly improve performance.

\subsection{Viewing \ours~from Other Perspectives}
In previous sections, we design \ours~from a compression perspective. 
Here, we further discuss it from the perspectives of \textit{Memory} and \textit{KV Cache Compression}, where KV Cache can be viewed as a form of LLM work memory.

In Memory perspective, \ours's workflow can be summarized as follows: it first performs autoregressive reasoning, then stores key information from the reasoning process as memory (memory), and continues reasoning based on the memorized content. 
Thus, the information in the cache tokens acts as a compact memory, though it is only effective for the current LLM and lacks transferability.

In KV Cache Compression perspective,
unlike methods such as H2O~\citep{nips23_h2o}, which rely on manually designed eviction policy to select important tokens, \ours~merges previous tokens in a continuous space, \textit{ceating} new representations. 
The content and manner of merging are autonomously determined by the LLM, rather than being a discrete selection process.



\begin{figure*}[!htbp]
\centering
\scalebox{1}{
\begin{tcolorbox}
\textbf{System Prompt:}

Below is a question. Please think through it step by step, and then provide the final answer. If options are provided, please select the correct one. 

\#\# Output format:\\
Use ``<THOUGHT>...</THOUGHT>'' to outline your reasoning process, and enclose the final answer in `\textbackslash boxed\{\}`.\\
\\
\#\# Example 1:\\
Question: \\
What is 2 + 3?\\
Output:\\
<THOUGHT>First, I recognize that this is a simple addition problem. Adding 2 and 3 together gives 5.</THOUGHT>\\
Therefore, the final answer is \textbackslash boxed\{5\}.\\
\\
\#\# Example 2:\\
Question: \\
What is 2 + 3?\\
A. 4\\
B. 5\\
C. 10\\
\\
Output:\\
<THOUGHT>First, I recognize that this is a simple addition problem. Adding 2 and 3 together gives 5.</THOUGHT>\\
Therefore, the final answer is \textbackslash boxed\{B\}.\\

\end{tcolorbox}
}

\caption{System prompt for \texttt{Qwen2.5-7B-Instruct} and \texttt{Llama3.1-8B-Instruct}.}
\label{prompt:system:base}
\end{figure*}

\begin{figure*}[!htbp]
\centering
\scalebox{1}{
\begin{tcolorbox}
\textbf{System Prompt:}

Your role as an assistant involves thoroughly exploring questions through a systematic long thinking process before providing the final precise and accurate solutions. This requires engaging in a comprehensive cycle of analysis, summarizing, exploration, reassessment, reflection, backtracing, and iteration to develop well-considered thinking process. Please structure your response into two main sections: Thought and Solution. In the Thought section, detail your reasoning process using the specified format: \texttt{<|begin\_of\_thought|>} \{thought with steps separated with `\textbackslash n\textbackslash n'\} \texttt{<|end\_of\_thought|>} Each step should include detailed considerations such as analisying questions, summarizing relevant findings, brainstorming new ideas, verifying the accuracy of the current steps, refining any errors, and revisiting previous steps. In the Solution section, based on various attempts, explorations, and reflections from the Thought section, systematically present the final solution that you deem correct. The solution should remain a logical, accurate, concise expression style and detail necessary step needed to reach the conclusion, formatted as follows: \texttt{<|begin\_of\_solution|>}  \{final formatted, precise, and clear solution\} \texttt{<|end\_of\_thought|>}  Now, try to solve the following question through the above guidelines:

\end{tcolorbox}
}

\caption{System prompt for \texttt{Vanilla}, \texttt{H2O}, \texttt{SepLLM}, \texttt{AnLLM}, and \texttt{\ours} for both Qwen-based model and Llama-based model.}
\label{prompt:system:vanilla}
\end{figure*}

\begin{figure*}[!htbp]
\centering
\scalebox{1}{
\begin{tcolorbox}
\textbf{GSM8K/MMLU/GPQA/BBH:}

Return your final response within \textbackslash boxed\{\}. \textcolor{violet}{\texttt{\{Question\}}}

\end{tcolorbox}
}

\caption{Task prompt for \texttt{Distill-R1}, \texttt{Vanilla}, \texttt{H2O}, \texttt{SepLLM}, \texttt{AnLLM}, and \texttt{\ours} for both Qwen-based model and Llama-based model.}
\label{prompt:task:r1}
\end{figure*}

\begin{figure*}[!htbp]
\centering
\scalebox{1}{
\begin{tcolorbox}
\textbf{GSM8K:}

Return your final response within \textbackslash boxed\{\}. \textcolor{violet}{\texttt{\{Question\}}}\\
\\
\textbf{MMLU:}

Please select the option that best answers the question. Return your final response within \textbackslash boxed\{\}. Here are the Question:\\
\textcolor{violet}{\texttt{\{Question\}}}\\

\textbf{GPQA:}

Given a question, please select the option that best answers it. Return your final response within \textbackslash boxed\{\}.
\textcolor{violet}{\texttt{\{Question\}}}\\

\textbf{BBH:}

Return your final response within \textbackslash boxed\{\}. If options are provided, please select the correct one. \textcolor{violet}{\texttt{\{Question\}}}

\end{tcolorbox}
}

\caption{Task prompt for \texttt{Qwen2.5-7B-Instruct} and \texttt{Llama3.1-8B-Instruct}.}
\label{prompt:task:base}
\end{figure*}

\begin{figure*}[!htbp]
\centering
\scalebox{1}{
\begin{tcolorbox}
\textbf{System Prompt:}

See Figure~\ref{prompt:system:vanilla}
\\

\textbf{Input:}

Return your final response within \textbackslash boxed\{\}. The average of the numbers 23 and $x$ is 27. What is the positive difference between 23 and $x$? (Recall that the average of two numbers is their sum divided by 2).
\\

\textbf{Output:}

\texttt{<|begin\_of\_thought|>}
Okay, let's see. The problem says that the average of the numbers 23 and x is 27. We need to find the positive difference between 23 and x. Hmm, right. Since the average of two numbers is their sum divided by 2, I can start by setting up the equation for the average.\textcolor{red}{\texttt{<w>}}So, the average of 23 and x is (23 + x)/2. According to the problem, this average equals 27. So I can write the equation:\textcolor{red}{\texttt{<w>}}(23 + x)/2 = 27\textcolor{red}{\texttt{<w>}}To find x, I should solve this equation. Let me multiply both sides by 2 to get rid of the denominator. That gives:\textcolor{red}{\texttt{<w>}}23 + x = 54\textcolor{red}{\texttt{<w>}}Then, subtract 23 from both sides to isolate x:\textcolor{red}{\texttt{<w>}}x = 54 - 23\textcolor{red}{\texttt{<w>}}Calculating that, 54 minus 23 is 31. So x is 31. Now, the question asks for the positive difference between 23 and x. The positive difference would be the absolute value of 23 - x. Since x is 31, that's |23 - 31|, which is | -8 |, so 8. Alternatively, it could also be |31 - 23|, which is the same result, 8. Either way, the positive difference is 8.\textcolor{red}{\texttt{<w>}}Wait, let me check my steps again to make sure I didn't make a mistake. Starting with the average: (23 + x)/2 = 27. Multiply both sides by 2: 23 + x = 54. Subtract 23: x = 31. Difference between 23 and 31 is 8. That seems right. Yeah, I think that's correct. No issues here.\textcolor{red}{\texttt{<w>}}
\texttt{<|end\_of\_thought|>}

\texttt{<|begin\_of\_solution|>}
\textcolor{red}{\texttt{<w>}}The average of 23 and \( x \) is given by \(\frac{23 + x}{2} = 27\). Multiplying both sides by 2:  
\[
23 + x = 54
\]  
Subtracting 23 from both sides:  
\[
x = 54 - 23 = 31
\]  
The positive difference between 23 and \( x \) is:  
\[
|31 - 23| = 8
\]  \textcolor{red}{\texttt{<w>}}\(\boxed{8}\)
\texttt{<|end\_of\_solution|>}

\end{tcolorbox}
}

\caption{A training case. \texttt{<w>} is a compression trigger.}
\label{prompt:case:train}
\end{figure*}
\begin{figure*}[!t]
\centering
\scalebox{0.9}{
\begin{tcolorbox}
\textbf{Question:}\\
Boris owns a chocolate factory. He produces 50,000 bars of chocolate each month. Boris produces 8,000 bars of chocolate the first week. The second week, Boris only produces half as much as the first week. But, the third week, Boris produces three times as much as the first week. How much does he produce the fourth week?\\ \\
\textbf{Model Output:}\\
\texttt{<|begin\_of\_thought|>} \\
Okay, let's see. Boris has a chocolate factory that produces 50,000 bars each month. The question is asking how much he produces in the fourth week. Hmm, let me break this down step by step.
<|splitter|>\\ 
\\First, the total monthly production is 50,000 bars. The month is divided into four weeks, right? So the production for each week adds up to 50,000. The problem gives the production for the first three weeks and asks for the fourth week. Let me note down the given information.
<|splitter|>\\
\\First week: 8,000 bars. That's straightforward. Then the second week, he produces half as much as the first week. Half of 8,000 is 4,000. So second week is 4,000 bars. Then the third week, he produces three times as much as the first week. Three times 8,000 is 24,000. So third week is 24,000 bars.
<|splitter|>\\
\\Now, to find the fourth week's production, I need to subtract the sum of the first three weeks from the total monthly production. Let me add up the first three weeks: 8,000 + 4,000 + 24,000. Let's do the math step by step. 8,000 plus 4,000 is 12,000. Then 12,000 plus 24,000 is 36,000. So the total for the first three weeks is 36,000 bars.
<|splitter|>\\
\\The total monthly production is 50,000 bars. Subtracting the first three weeks from the total gives the fourth week's production. So 50,000 minus 36,000 equals 14,000 bars. Therefore, Boris produces 14,000 bars in the fourth week.
<|splitter|>\\ 
\\Wait a second, let me double-check my calculations to make sure I didn't make a mistake. First week: 8,000. Second week: half of 8,000 is indeed 4,000. Third week: three times 8,000 is 24,000. Adding those: 8,000 + 4,000 = 12,000; 12,000 + 24,000 = 36,000. Total monthly is 50,000. So 50,000 - 36,000 = 14,000. Yep, that seems right. I don't think I missed anything here. The problem states the production for each week as fractions or multiples of the first week, so the calculations are straightforward. No complicated formulas needed. Just basic arithmetic. So the answer should be 14,000 bars in the fourth week.
<|splitter|>\\
\texttt{<|end\_of\_thought|>}\\
\\ \texttt{<|begin\_of\_solution|>}
\\Boris's weekly chocolate production is calculated as follows:
<|splitter|>\\
\\- \textbf{First week}: 8,000 bars  \\
- \textbf{Second week}: \( \frac{8,000}{2} = 4,000 \) bars \\ 
- \textbf{Third week}: \( 8,000 \times 3 = 24,000 \) bars  <|splitter|>\\
\\Total for the first three weeks:  \\
$4,000 + 4,000 + 24,000 = 32,000 \text{ bars}$
<|splitter|> \\
\\Subtracting this from the monthly total: \\ 
$50,000 - 32,000 = 18,000 \text{ bars}$
<|splitter|>\\
\\Thus, Boris produces \(\boxed{18000}\) bars in the fourth week. <|splitter|>\\
\texttt{<|begin\_of\_solution|>}

\end{tcolorbox}
}

\caption{Bad Case. \texttt{<|splitter|>} is equal to \texttt{<w>} in Figure~\ref{prompt:case:train}.}
\label{prompt:case}
\end{figure*}

\subsection{Why \ours~generates more tokens with smaller cache size?}
As shown in Figure~\ref{fig:exp:efficient}(e-f), we find that \ours~generates more tokens with smaller cache size.
We examined outputs under different cache sizes and found that when the cache size is small, the model tends to repeat previous content more often. 
We believe this is because smaller cache sizes lead to greater information loss during compression, prompting the model to regenerate earlier content more frequently to retain as much information as possible.

\subsection{Comparison with Implicit CoT Works}
\begin{table*}[h!]
\small
\centering
\begin{tabular}{lcccc}
\toprule
\textbf{Method} & \textbf{Memory Optimization} & \textbf{Training Cost} & \textbf{Interpretability} & \textbf{Generalization} \\
\midrule
System-1.5~\cite{arxiv25_system_1_5} & Significant & High  & Weak & Weak \\
SoftThinking~\cite{arxiv25_soft_thinking}  & Limited    & None  & Weak & Good \\
\ours & Significant & Low   & Good & Good \\
\bottomrule
\end{tabular}
\caption{Comparison of different methods.}
\label{tab:app:comparison}
\end{table*}

Both our work and implicit CoT methods operate in continuous spaces.
While implicit CoT performs reasoning entirely in continuous space, \ours~employs a hybrid approach combining continuous and discrete space reasoning. 
The overview differences are shown in Table~\ref{tab:app:comparison}.
Below we clarify the key differences:
\begin{itemize}
    \item \textbf{Reasoning Acceleration Mechanism}.
    \emph{Implicit CoT} methods (e.g., System-1.5~\cite{arxiv25_system_1_5}, SoftThinking~\cite{arxiv25_soft_thinking}) accelerate reasoning by reducing generation steps through continuous token representations. 
    The entire reasoning process depends on full context.
    \emph{\ours} accelerates inference by reducing the number of historical tokens needed for generation, without requiring full context dependence.

    \item \textbf{Training Approach.}
    Implicit CoT typically requires complex, multi-stage training with significant overhead (e.g., System-1.5's two-phase training~\cite{arxiv25_system_1_5} or Coconut's curriculum learning~\citep{arxiv24_coconut}). Some methods~\cite{arxiv25_soft_thinking, arxiv24_ccot} even require architecture modifications.
    \ours~uses standard SFT with modified attention masks, requiring neither specialized training data nor architectural changes.

    \item \textbf{Interpretability and Generalization}.
    Current implicit CoT methods face interpretability challenges (due to continuous reasoning) and limited out-of-domain generalization (except SoftThinking~\cite{arxiv25_soft_thinking}).
    \ours~maintains discrete tokens for better interpretability and shows promising out-of-domain generalization in our experiments, as shown in Table~\ref{tab:app:comparison}.

\end{itemize}


\end{document}